\documentclass[sigconf,balance=false]{acmart}

\usepackage{booktabs}
\usepackage{siunitx}
\usepackage{xcolor}
\usepackage{listings}
\usepackage[htt]{hyphenat}

\usepackage{array}
\newcolumntype{L}[1]{>{\raggedright\arraybackslash}p{#1}}

\newif\iffigpdf
\figpdftrue        

\newif\iffigdraft
\figdraftfalse        

\iffigpdf
  \usepackage{graphicx}
\else
  \iffigdraft\else
    \usepackage{tikz}
    \usetikzlibrary{arrows.meta,positioning,fit,backgrounds,calc}
  \fi
\fi

\newcommand{\missingfigure}[2]{%
  \begin{figure}[t]\centering
  \fbox{\parbox{0.90\columnwidth}{\centering\small
  Figure artwork from the original project is not included in this file.}}
  \caption{#1}\label{#2}\end{figure}}
\IfFileExists{figures.tex}{

\newcommand{\figDesign}{%
\begin{figure*}[t]
\centering
\includegraphics[width=\textwidth]{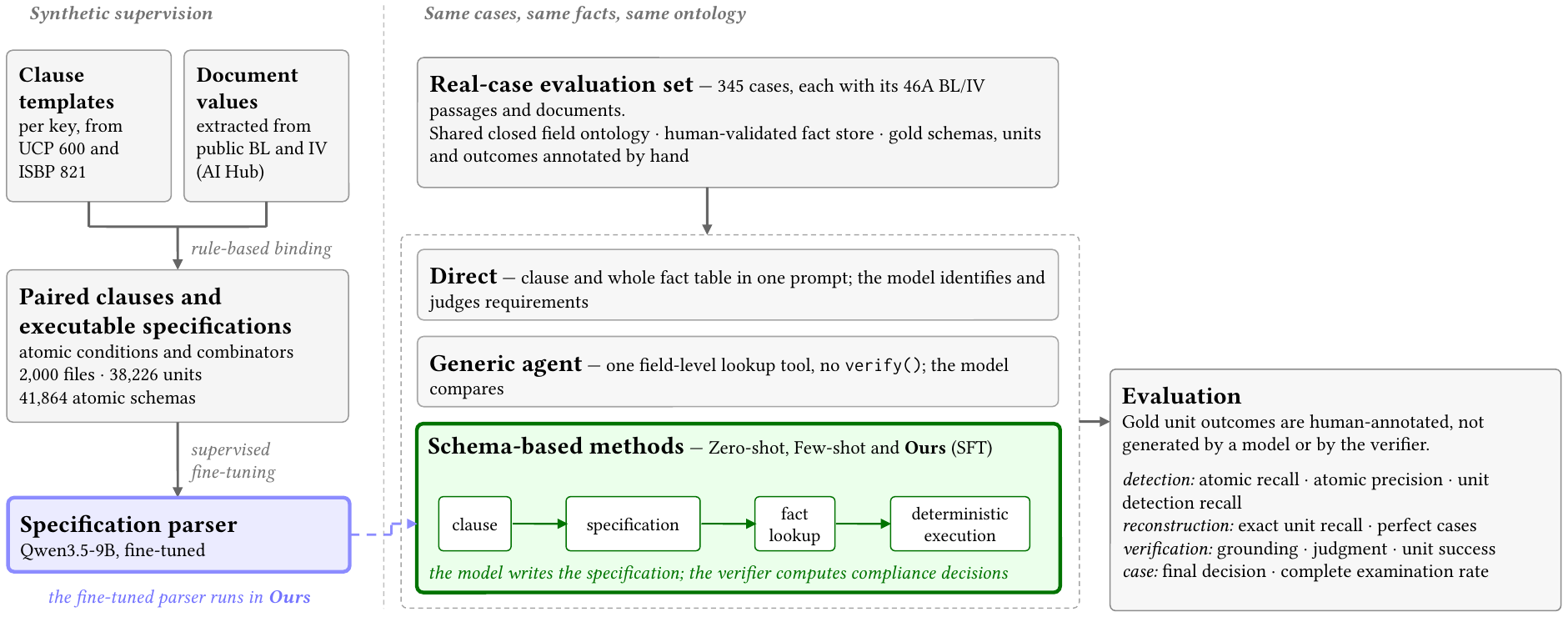}
\caption{Study design. Synthetic pairs train the parser (left); every
method is evaluated on the same real cases, facts and ontology against
human-annotated gold outcomes (right).}
\Description{A two-part diagram. On the left, clause templates and
document values extracted from public bills of lading and commercial
invoices are combined by rule-based binding into paired clauses and
executable specifications, which train a specification parser by
supervised fine-tuning. On the right, a box describing the real
benchmark of 345 cases with a shared field ontology and a
human-validated fact store feeds three method boxes: a direct model
given the whole fact table in one prompt, a generic agent with a
field-level lookup tool, and the schema-based methods --- zero-shot,
few-shot and the proposed fine-tuned parser --- which go from clause to
specification to fact lookup to deterministic execution, the model
writing the specification and the verifier computing the compliance
decisions. A dashed arrow runs from the fine-tuned parser into the
schema-based methods. All methods feed one evaluation box that states
the gold outcomes are human-annotated and lists the detection,
reconstruction, verification and case-level metrics.}
\label{fig:design}
\end{figure*}}

\newcommand{\figIncomplete}{%
\begin{figure*}[t]
\centering
\includegraphics[width=\textwidth]{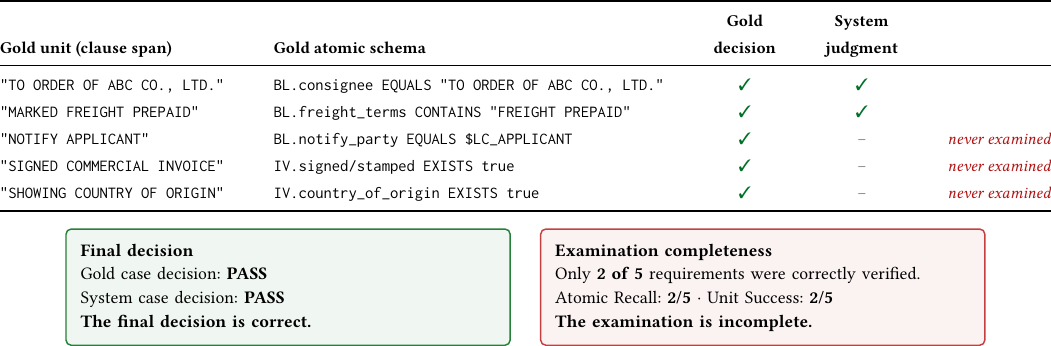}
\caption{Illustrative example with five atomic, evaluable units.
All five gold unit decisions are PASS, but the system correctly verifies
only two and returns PASS for the case. Its final decision is correct,
while Atomic Recall and Unit Success are both $2/5$ and the examination
is incomplete.}
\Description{Six clause spans are paired with their gold atomic schemas,
gold decisions and system judgments. All gold decisions are PASS.
The first three system judgments are PASS; the remaining three units
are not examined. Two summary boxes contrast a correct final PASS
decision with incomplete examination: only three of six requirements
are correctly verified.}
\label{fig:incomplete}
\end{figure*}}

\newcommand{\figLayers}{%
\begin{figure}[t]
\centering
\includegraphics[width=\columnwidth]{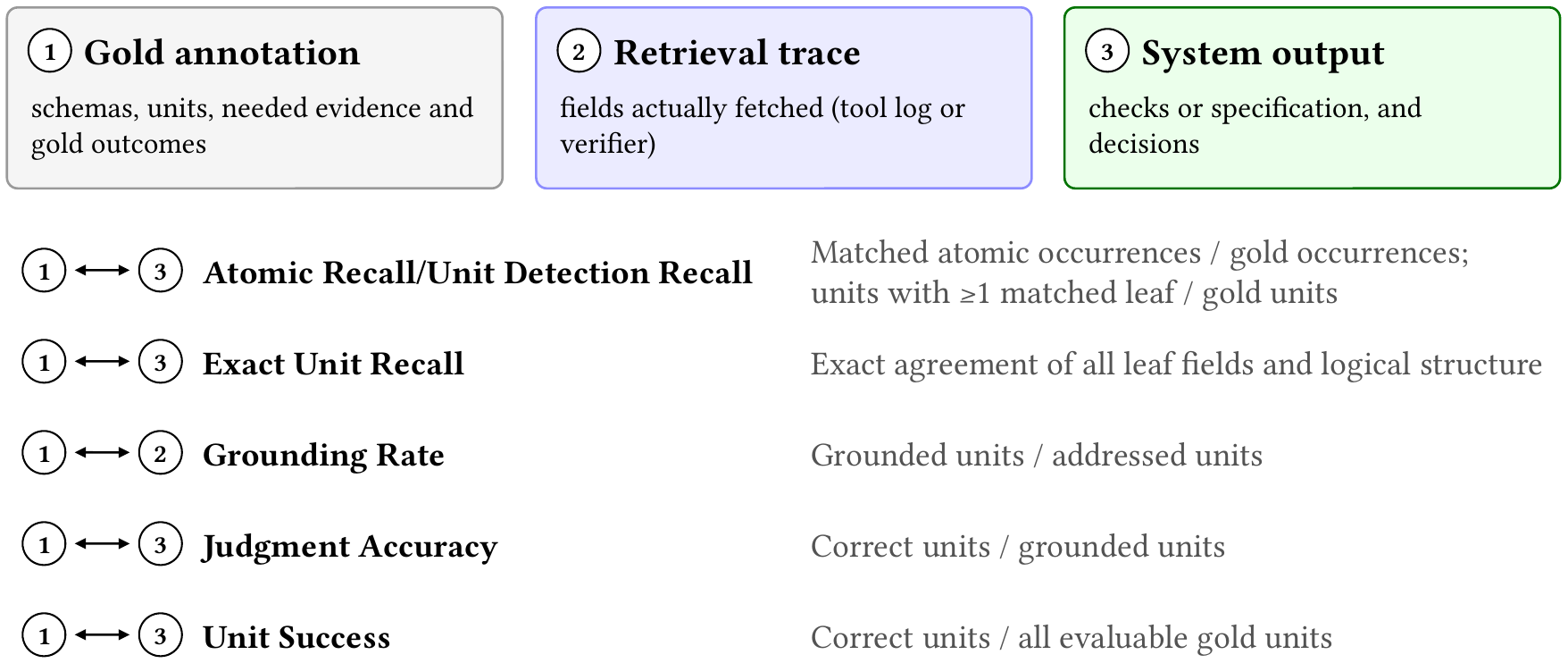}
\caption{The observable layers each metric compares. For schema-based
methods the trace is the verifier's retrieval; Direct has none, so its
grounding is not applicable. Arrows indicate the primary comparison;
retrieval traces determine grounding eligibility for Judgment Accuracy
and Unit Success.}
\Description{Three boxes across the top: the gold annotation, which
gives the atomic schemas, their grouping into units, the evidence each
unit needs and each unit's outcome; the retrieval trace, which gives the
fields actually accessed; and the system output, which gives the checks
or specification a method emits and its decisions. Below them, five
rows name a metric, the pair of layers it compares and how it is
computed: Atomic Recall and Unit Detection Recall, and Exact Unit
Recall, compare the gold annotation with the system output; Grounding
Rate compares the gold annotation with the retrieval trace; Judgment
Accuracy and Unit Success compare the gold annotation with the system
output.}
\label{fig:layers}
\end{figure}}

\newcommand{\figKeys}{%
\begin{figure}[t]
\centering
\includegraphics[width=\columnwidth]{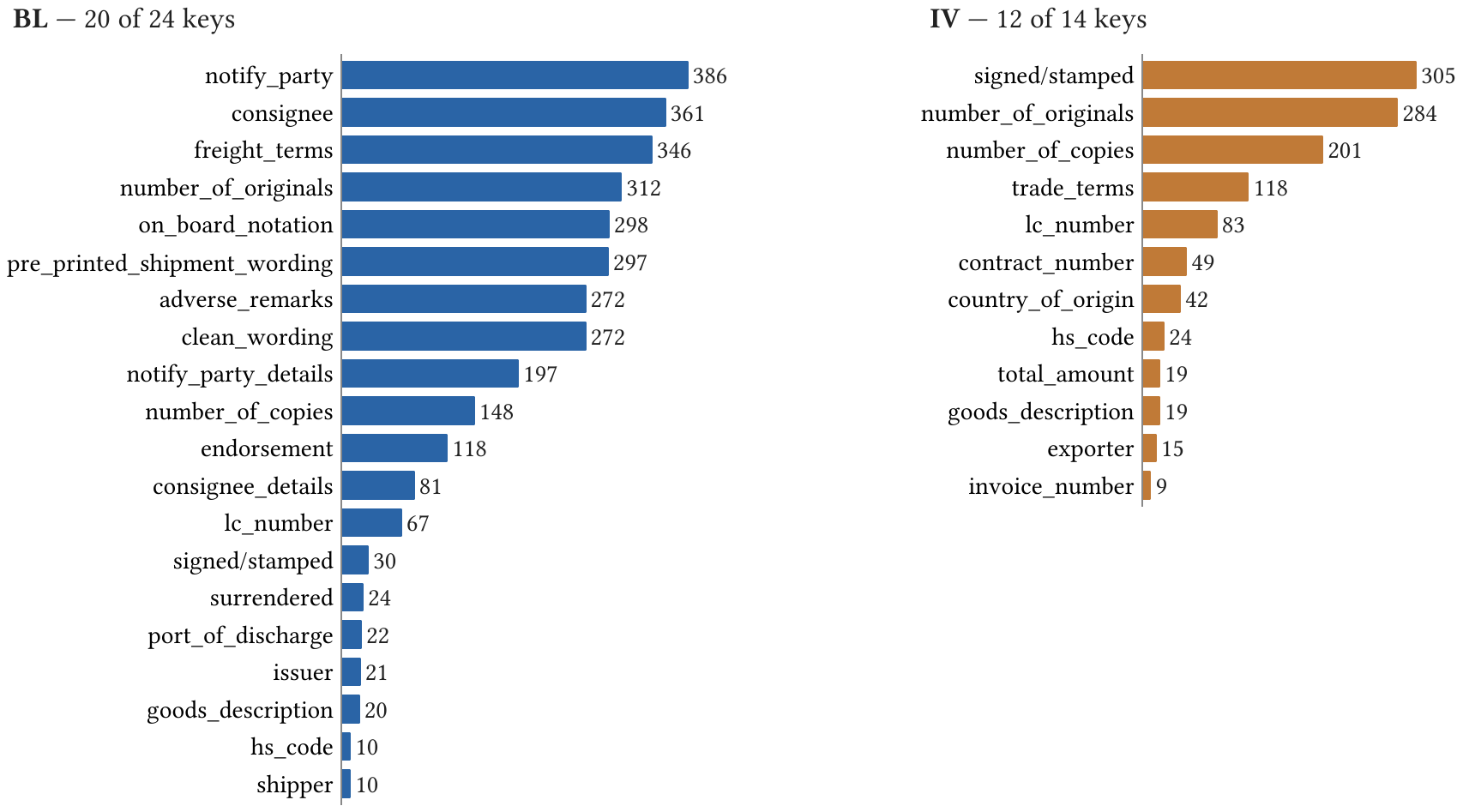}
\caption{Gold atomic key frequencies in 345 cases. The 20 most frequent
BL keys and 12 most frequent IV keys account for 4,460 of the 4,493
key-covered occurrences; the 378 key-level out-of-ontology requirements are not
shown.}
\Description{Two side-by-side panels of horizontal bars: 20 of 24 BL
keys and 12 of 14 IV keys. The BL panel ranges from notify party at
386 occurrences to hs code and shipper at 10 each. The IV panel ranges
from signed/stamped at 305 to invoice number at 9.}
\label{fig:keys}
\end{figure}}
}{%
  \newcommand{\figIncomplete}{\missingfigure{A correct final decision without a complete examination.}{fig:incomplete}}
  \newcommand{\figDesign}{\missingfigure{Specification-guided verification architecture.}{fig:design}}
  \newcommand{\figLayers}{\missingfigure{Requirement detection, executable structure reconstruction, and verification are evaluated separately.}{fig:layers}}
  \newcommand{\figKeys}{\missingfigure{Gold atomic key frequencies in 345 cases.}{fig:keys}}
}

\lstdefinestyle{py}{
  language=Python,
  basicstyle=\ttfamily\scriptsize,
  keywordstyle=\color{blue!60!black},
  commentstyle=\color{gray},
  stringstyle=\color{green!40!black},
  showstringspaces=false,
  columns=fullflexible,
  breaklines=true,
  frame=single,
  framesep=3pt,
  rulecolor=\color{black!30},
  aboveskip=4pt,
  belowskip=2pt,
}

\begin{document}
\setcopyright{none}
\renewcommand\footnotetextcopyrightpermission[1]{}
\acmConference[AI4F @ ACM ICAIF '26]{The 3rd Workshop on LLMs and Generative AI for Finance (AI4F), co-located with the 7th ACM International Conference on AI in Finance}{November 14--17, 2026}{Milan, Italy}

\title[From Letter-of-Credit Requirements to Executable Specifications]{From Letter-of-Credit Requirements to Executable Specifications: Evaluating Verification Accuracy and Completeness}

\author{Ji Sub Choi}
\authornote{These authors contributed equally to this research.}
\affiliation{%
  \institution{Hana Institute of Technology, Hana TI}
  \city{Seoul}
  \country{South Korea}}
\email{jisubchoi@hanafn.com}

\author{Whan Choi}
\authornotemark[1]
\affiliation{%
  \institution{Hana Institute of Technology, Hana TI}
  \city{Seoul}
  \country{South Korea}}
\email{wchoi2152@hanafn.com}

\author{Seo Yeon Gu}
\authornotemark[1]
\affiliation{%
  \institution{Hana Institute of Technology, Hana TI}
  \city{Seoul}
  \country{South Korea}}
\email{sygu@hanafn.com}

\author{Suhyun Kim}
\affiliation{%
  \institution{Hana Institute of Technology, Hana TI}
  \city{Seoul}
  \country{South Korea}}
\email{suhyun\_kim@hanafn.com}

\author{Inyong Yun}
\affiliation{%
  \institution{Hana Institute of Technology, Hana TI}
  \city{Seoul}
  \country{South Korea}}
\email{iyyun@hanafn.com}

\renewcommand{\shortauthors}{Choi et al.}

\begin{abstract}
In letter-of-credit (LC) document examination, a correct final decision does not show that every requirement was examined. We therefore evaluate examination requirement by requirement and make each requirement executable. A language model converts the bill of lading (BL) and commercial invoice (IV) requirements of a credit into atomic conditions and logical structures, and a deterministic verifier retrieves the named evidence and computes each outcome, so every check can be compared with a gold encoding and traced to logged evidence. Because LC records are confidential, our on-premises 9B parser is fine-tuned only on synthetic pairs built from templates and human-reviewed public document values. On 345 real LC cases, final-decision accuracy proves misleading: a constant PASS rule has the highest raw accuracy while examining nothing. Although it never sees a real record during training, the fine-tuned parser detects 93\% of the atomic requirements its ontology covers and reconstructs 86\% of the covered requirements exactly. It verifies 82\% of the evaluable requirements correctly, against 54\% for a tool-using agent, and leads the agent on both satisfied requirements and discrepancies. It also completely examines 1.8 times as many cases as few-shot prompting of the same model, while direct prediction and the agent complete none. Executing the gold encodings with the same verifier shows that some of the remaining errors, missed discrepancies included, come from the operators and normalization, which a better parser alone cannot fix. Every rejection comes with a trace of the units and evidence behind it, which suits examiner-in-the-loop review.
\end{abstract}

\maketitle
\begingroup
\renewcommand\thefootnote{}
\footnotetext{Accepted at the 3rd Workshop on LLMs and Generative AI for Finance (AI4F) at ACM ICAIF '26, Milan, Italy (non-archival).}
\endgroup

\section{Introduction}
\label{sec:intro}

Under a documentary credit, or letter of credit (LC), the issuing bank must honour a presentation of documents that complies with the credit's terms. Banks decide compliance on the documents alone (UCP 600 Art.~14(a) \cite{ucp600}) and within five banking days (Art.~14(b)). The requirements are free text and span several documents. Language models are a natural candidate for this work, and the usual evaluation compares their final PASS/FAIL decision with a human examiner's. We argue that this is not enough: the examination itself must be evaluated, requirement by requirement.

A single clause can set several conditions on the BL and the IV, such as parties, required wording, dates, and references to the credit. The final PASS/FAIL label compresses all of them into one decision, so a system can skip a satisfied requirement and still return the correct label (Figure~\ref{fig:incomplete}). This gap matters for both kinds of presentation. For a compliant presentation, the bank must establish compliance by examining the documents (Art.~14(a)); a correct PASS with an unexamined requirement has not established it. For a discrepant presentation, a bank that refuses must send a single notice that states each discrepancy on which it refuses (Art.~16(c)), and a bank that does not act in this way cannot then claim that the documents do not comply (Art.~16(f)). A discrepancy left out of the notice therefore cannot be relied on later, so in practice the examiner must find and list every discrepancy that exists. A correct FAIL label is not enough either, because it does not show which discrepancies were found.
\figIncomplete

We therefore evaluate examination at four levels, each answering one question. (i) Did the system identify the requirements? (ii) Did it reconstruct their conditions and logical structure? (iii) Do its evidence-based outcomes agree with human judgments? (iv) Did it examine every evaluable requirement and reach the correct decision? The levels differ because an outcome can be right without a faithful reconstruction. For A OR B, a passing A decides the unit even if B is never reconstructed, and a condition with the wrong operator can still give the right label on a given presentation.

To evaluate levels (ii) and (iii), generated requirements must be comparable with gold annotations, and every outcome must be traceable to evidence. We therefore make the requirements executable (Figure~\ref{fig:design}). A parser converts the clause into a \emph{specification}: atomic conditions (atomic schemas) grouped into top-level verification units. A deterministic verifier retrieves the facts that the specification names and computes the outcomes. The model decides what to check, and the verifier decides whether each check passes. Executability serves evaluation and practice alike. An evaluator can compare each check with a gold encoding, and can execute the gold encodings themselves to see which errors the representation introduces. An examiner can see which unit and which evidence led to each decision. Specifications must still be generated without exposing confidential data: LC records cannot be sent to external model services and are rarely available for training. We therefore fine-tune a 9B parser that runs on premises, using only synthetic pairs built from clause templates and human-reviewed public document values, and we use the real cases only for evaluation.
\figDesign

We study three research questions. \textbf{RQ1:} How often do correct final decisions come with incomplete examination, and are correct refusals based on the annotated discrepancies? \textbf{RQ2:} How do specification-guided and model-based systems compare in coverage, unit outcomes, and completeness? \textbf{RQ3:} How does a parser fine-tuned on synthetic specifications compare with few-shot prompting of the same base model? Our contributions are:

\begin{itemize}
\item \textbf{A requirement-level evaluation of LC examination on 345 real cases.} It measures detection, exact reconstruction, unit outcomes, and complete examination separately. Any system that reports its checks by document and field can be scored, so other institutions can apply the protocol once they have annotated their requirements and built a fact store. To our knowledge, it is the first such evaluation. It shows that final-decision accuracy hides incomplete examination, even for correct refusals.
\item \textbf{An executable representation that makes examination inspectable.} Six reviewable operators, logical connectives, and credit-field references let a deterministic verifier compute every outcome from logged evidence, so each check is reproducible and auditable. Because generated specifications share the format of the gold encodings, omissions and operator errors can be measured, and executing the gold encodings with the same verifier separates generation errors from errors that the representation itself introduces, which an independent expert review traces to operator expressiveness and normalization.
\item \textbf{Synthetic supervision that makes specification generation feasible without confidential LC records.} A 9B parser that runs on premises and is trained only on template-generated pairs transfers to real credits. It reaches the highest Exact Unit Recall among the schema-based methods and the highest Complete Examination Rate among all evaluated systems, while direct prediction and a tool-using agent examine no case completely with the same 9B model. We release the generation code and the public-source templates.
\end{itemize}

\section{Background and Problem Formulation}
\label{sec:background}

A credit is usually issued as a SWIFT MT700 message. Its field 46A, Documents Required, lists the documents and their conditions in free text, for example BLs to the order of the issuing bank marked freight prepaid, and a signed IV showing specified details. Requirements are not grouped by document. Annotators selected the field-46A passages on the BL and IV and joined them into one input clause without regrouping, so assigning requirements to documents is part of the task. Requirements may also refer to other credit fields: ``to the order of the issuing bank'' requires resolving the issuing bank before comparing it with the consignee.

A case $c$ consists of this clause and a fact store of human-validated values from the presented documents and the credit, over a fixed field ontology. An \emph{atomic schema} is a tuple (document, key, operator, expected value), which encodes one atomic requirement and forms a leaf of a unit. The gold verification units $U(c)$ are the top-level requirements being judged. Each unit is a single schema or a tree of schemas joined by AND, OR, or AND/OR, a form common in LC drafting that is executed as OR (Section~\ref{sec:schemas}). Annotators label each atomic schema PASS or FAIL, and a unit's outcome follows from these labels and its logical structure. For example, ``NOTIFY A OR B'' is one unit with two atomic schemas. The gold schema collection $S(c)$ is the multiset of atomic schemas in $U(c)$, and a specification is the set of units that a parser generates for a clause. Detection and reconstruction are scored on all units whose keys belong to the ontology (key-covered units; Section~\ref{sec:benchmark}). Verification and case-level results use the evaluable scope of Section~\ref{sec:eligibility}, where the gold case decision is the conjunction of the evaluable unit outcomes.

\section{Related Work}
\label{sec:related}

\textbf{Process-level evaluation.} A correct output can come from a flawed process. Models can be right for the wrong reasons \cite{mccoy2019right}, stated reasoning need not reflect what drove a prediction \cite{turpin2023unfaithful}, and language models can claim conclusions from formal reasoning they never executed \cite{knowyourlimits}. Agent benchmarks such as $\tau$-bench \cite{yao2024taubench} score the final state an agent produces in addition to its reply, and AuditAgent \cite{bai2025auditagent} scores the issues and evidence an agent recovers in fraud audits. None of these evaluates whether each requirement of a rule set was examined, which LC practice requires.

\textbf{Executable checks.} When the model itself judges, as in our Direct and Agent baselines, identical inputs can yield different decisions \cite{outputdrift}. A tool-using agent \cite{yao2023react} chooses what to retrieve and may never consult the relevant evidence. Program-aided reasoning hands the computation to an interpreter \cite{gao2023pal}. In financial compliance, Compliance-to-Code \cite{compliance2code} and a neuro-symbolic system \cite{neurosymbolic} generate free-form checking code or SMT constraints, whose faithfulness must be checked program by program, and LAVA \cite{shu2025lava} executes arithmetic rules but leaves symbolic rules to the model. Our verifier works with a closed set of six operators that examiners can review, and because generated specifications can be compared with gold encodings, omissions and operator errors become measurable. Following the distinction between exact match and execution accuracy in text-to-SQL \cite{yu2018spider,zhong2020testsuite}, we report Exact Unit Recall and Unit Success separately.

\textbf{LC examination and training data.} Early LC automation parsed free-text applications with case frames \cite{dhar1986lc}. Later work compared AI-based checking with human checkers \cite{khalil2022ieom} and optimized how human verification is allocated \cite{khalil2025lc}, reporting error counts and processing times. A recent roadmap reviews bank deployments of commercial checking products and notes the lack of quantitative validation \cite{khalil2025roadmap}. To our knowledge, no academic study evaluates LC examination at the requirement level on real cases. Semantic parsers have been trained on grammar- or template-generated pairs, usually paraphrased by crowd workers \cite{wang2015overnight}; our pairs are generated by templates alone, each clause together with its specification.

\section{Specification-Guided Verification}
\label{sec:method}

\subsection{Schemas and units}
\label{sec:schemas}

An atomic schema has a document, an ontology key, one of six operators (Table~\ref{tab:ops}), and, where needed, an expected value. The expected value is a literal or a credit-field reference such as \texttt{\$LC\_APPLICANT}, which is resolved at execution. EXISTS is encoded with the value \texttt{true} (\texttt{IV.country\_of\_origin EXISTS true} passes if the field has any value). NEGATION passes only when the field has no value, as when a BL must bear no adverse remarks. Composite units join atomic schemas or nested nodes with AND, OR, or AND/OR. AND/OR executes as OR but is kept separate so that the trace follows the clause wording. Each top-level requirement counts as one verification unit; nested logical groups remain part of that unit.

\begin{table}[t]
\centering
\small
\caption{The six supported operators. Requirements that need other operations cannot be expressed. NEGATION denotes field-value absence.}
\label{tab:ops}
\setlength{\tabcolsep}{4pt}
\begin{tabular}{@{}lll@{}}
\toprule
Operator & Test & Typical use \\
\midrule
\texttt{EQUALS}   & normalized equality    & parties, currency, enumerations \\
\texttt{CONTAINS} & normalized substring   & free text, marks, addresses \\
\texttt{EXISTS}   & field present          & data required to appear \\
\texttt{LTE}      & numeric or date $\leq$ & deadlines, ceilings \\
\texttt{GTE}      & numeric or date $\geq$ & counts, floors \\
\texttt{NEGATION} & value absent           & absence of adverse notations \\
\bottomrule
\end{tabular}
\end{table}

\subsection{Execution}
\label{sec:exec}

The verifier fetches the fields that the schemas name; for example, \texttt{BL.consignee EQUALS \$LC\_ISSUING\_BANK} reads \texttt{BL.consignee} and \texttt{LC.issuing\_bank}. It then resolves references, normalizes the values, applies the operators, and evaluates the unit logic recursively (Listing~\ref{lst:verify}). The prompt lists the 38 ontology keys. Requirements whose keys lie outside them, which we call out-of-ontology requirements, are reported in a separate out-of-ontology field that is neither executed nor evaluated here. Schemas that the parser places in this field although their key belongs to the ontology are moved into the specification before execution, for all three schema-based methods. Units on verification-ineligible keys (Section~\ref{sec:eligibility}) are skipped, and such branches are removed from composite units. The case decision is the conjunction of all executed units, including units without a gold counterpart. An absent value is returned as an empty string. Such a lookup still counts as retrieved evidence, which EXISTS and NEGATION test and which grounding requires (Section~\ref{sec:metrics}); a field that was never queried provides no evidence.

\textbf{Post-processing (normalization).} Only EQUALS, CONTAINS, LTE, and GTE normalize values. They apply the same rules to both compared values, for every schema-based method and for the gold encodings. EXISTS and NEGATION test the raw value. The rules were fixed before evaluation and are not used to produce the human labels. Text is converted to upper case and special characters are removed; synonyms are unified (CO to COMPANY, LTD to LIMITED); country names are standardized (REPUBLIC OF KOREA to KOREA); filler tokens such as THE are removed; and month names in dates are converted. For example, ``ABC CO., LTD.'' becomes ``ABC COMPANY LIMITED'', and ``TO THE ORDER OF X'' becomes ``TO ORDER OF X''. Because the order wording is kept, a consignment to the order of a bank and a straight consignment to that bank remain different values. Normalized comparison still departs from the ``not conflict'' standard of Art.~14(d), in both directions (Section~\ref{sec:analysis}).

For CLEAN, training targets and gold annotations use an OR of \texttt{BL.clean\_wording} and \texttt{BL.adverse\_remarks}, and reconstruction is scored against this encoding. For verification, the verifier follows Art.~27, under which a transport document is clean if it bears no notation of defective goods or packaging: it fails a CLEAN unit whenever an adverse remark is present and passes it otherwise. This rule replaces the executed result of any unit that contains \texttt{BL.clean\_wording} or \texttt{BL.adverse\_remarks}. For a unit generated as the OR encoding, it gives the same outcome as execution; it changes the outcome only when a CLEAN unit is generated in another form, and such a unit still counts as inexact in reconstruction.\footnote{No evaluation case contains an adverse remark. The gold OR encoding therefore gives the same outcomes, every unit containing either key passes for the schema-based methods, and the rule is untested on real failures.}

\begin{lstlisting}[style=py,float=t,caption={Generic deterministic execution after fact lookup; eq, contains, and compare normalize both values. The CLEAN rule is described in the text.},label={lst:verify}]
def verify_schema(s, facts, lc):
    observed = facts[s.doc][s.key]  # absent value is ""
    expected = resolve(s.expected_value, lc)
    if s.op == "EQUALS": return eq(observed, expected)
    if s.op == "CONTAINS": return contains(observed, expected)
    if s.op == "EXISTS": return observed != ""
    if s.op == "LTE": return compare(observed, expected) <= 0
    if s.op == "GTE": return compare(observed, expected) >= 0
    if s.op == "NEGATION": return observed == ""
    raise ValueError("Unsupported operator")

def verify_unit(u, facts, lc):
    if u.is_atomic(): return verify_schema(u.schema, facts, lc)
    out = [verify_unit(x, facts, lc) for x in u.children]
    if u.combinator == "AND": return all(out)
    if u.combinator in {"OR", "AND/OR"}: return any(out)
    raise ValueError("Unsupported combinator")
\end{lstlisting}

\paragraph{What execution guarantees.}
Every outcome at this layer is computed by the verifier from logged values and stays fixed for a given specification and fact store. This does not make the interpretation faithful: the parser can omit atomic schemas, choose a wrong operator or value, or group them wrongly. During evaluation, gold annotations expose these errors. In deployment, the trace shows what was checked but not what was missed.

\section{Synthetic Specification Supervision}
\label{sec:synthesis}

The parser is trained only on 2,000 generated files. Each file pairs a synthetic clause with its target specification; together they contain 38,226 units and 41,864 atomic schemas. Templates for each key follow examination patterns in UCP 600, ISBP 821 \cite{isbp821}, SWIFT MT700, and confidential bank drafting guidance. Their slots are filled with values that Qwen3.5-397B-A17B \cite{qwen35} extracts from public BLs and IVs in AI-Hub's OCR Data (Finance and Logistics).\footnote{\url{https://aihub.or.kr/aihubdata/data/view.do?dataSetSn=71301}} The three annotators (Section~\ref{sec:benchmark}) reviewed all extracted field values used for generation, in rotation. Rule-based binding inserts the same value into the clause and the target, so each pair is consistent by construction; OR, AND, AND/OR, and nesting also carry over from template to target. The pairs teach the parser to interpret requirements; judging compliance is left to the verifier. The parser never sees a real clause during training, so the evaluation tests transfer from synthetic to real requirement text. The two also differ in density: a synthetic file has 19.1 key-covered units on average, and a real case 11.2.

The template authors overlap with the annotators of the evaluation cases. The templates were finished before annotation and built from the sources above, not from evaluation cases, and the generator, prompts, demonstrations, and hyperparameters were fixed without using the evaluation cases. Training targets and gold annotations nonetheless share encoding conventions; Section~\ref{sec:rq3} discusses their effect. We release the generation code and the public-source templates.\footnote{https://github.com/Ji-S/LC-Spec-Synth}

We fine-tune Qwen3.5-9B \cite{qwen35} with LoRA \cite{hu2022lora} (rank 64, alpha 128, learning rate $2\times10^{-4}$, batch size 4), first for one epoch on 1,500 files whose party freight and port requirements contain no logical connectives (CLEAN and ON BOARD always use OR), and then for one epoch on 500 files with such connectives, so that the parser learns atomic schemas before their composition. No validation set was used, and we take the final checkpoint. Training and inference use the same three-example prompt as the Few-shot baseline.

\section{Real-Case Evaluation Set}
\label{sec:benchmark}

\subsection{Cases and key coverage}

The evaluation set has 345 real cases from one institution. Each case contains the field-46A requirements for the BL and IV, the presented BL and IV, and the relevant credit fields. Annotation yields 4,871 atomic requirements. The ontology of 38 keys (24 BL, 14 IV) was fixed before annotation, based on UCP 600 and practitioner checklists. Of the requirements, 4,493 (92.24\%) map to these keys, are encoded as atomic schemas, and form 3,848 gold units (Figure~\ref{fig:keys}). The other 378 are out-of-ontology requirements, and we exclude them. Being key-covered does not make a requirement expressible: a requirement that an HS code have six digits maps to a key but needs an operator we lack, and it stays in the evaluation. Most composite units are ORs; nested AND and AND/OR units also occur.
\figKeys

\subsection{Annotation and facts}

Three in-house annotators experienced in trade-finance examination defined the conventions together. They decomposed each requirement into atomic schemas with its logical structure, judged each atomic schema PASS or FAIL against the documents, cross-reviewed one another's work in rotation, and settled disagreements by reference to UCP 600. The outcomes are the annotators' own judgments and were never produced by the verifier. Cross-review checks a label that already exists, so it gives no independent second label from which inter-annotator agreement could be computed. As a separate check, two finance-domain experts outside the annotation team reviewed all 182 units on which the executed gold encodings disagree with the labels and found no label error (Section~\ref{sec:analysis}). Every method reads the same human-validated facts instead of document images, which keeps perception errors out of the comparison. The fact store lists all 38 keys for every case, empty where absent, so the full table given to Direct does not reveal which requirements a clause contains.

\subsection{Verification eligibility}
\label{sec:eligibility}

Three keys need evidence that the records do not contain. Endorsements are normally on the back of the BL, but only document faces were collected. Whether a full set of originals or the required copies was presented (\texttt{number\_of\_originals}, \texttt{number\_of\_copies}) cannot be read from one document face. These atomic schemas remain in detection and reconstruction, but a unit made only of them is not evaluable. In a composite unit, such branches are removed, and the unit's gold outcome is recomputed from the labels of the remaining atomic schemas. For example, ``to order of the issuing bank or to order of the shipper and blank endorsed'' is verified on its consignee schemas only, and such a reduced unit can be weaker than the original. Eligibility is decided by key alone and is the same for every method and outcome.

This leaves 2,817 of the 3,848 units (73.21\%) for verification: 2,731 PASS (96.95\%) and 86 FAIL. The gold case decision is the conjunction of these unit outcomes, giving 281 PASS and 64 FAIL cases. These shares describe a scope that omits field 47A, the other documents, and endorsement and set-composition conditions; they are not discrepancy rates of whole presentations. For the same reason, we derive the gold decision from the annotations: the bank's recorded decision covers the whole presentation and cannot be compared with a decision on this scope.

\section{Experimental Setup}
\label{sec:setup}

\subsection{Methods}
\label{sec:methods}

All methods run Qwen3.5-9B \cite{qwen35} on premises with the same cases, facts, and ontology (Table~\ref{tab:methods}), and none sees gold mappings from clauses to schemas or fields.

\begin{table}[t]
\centering
\small
\caption{Compared methods. Schema-based methods share a deterministic verifier. Mode: thinking or non-thinking generation.}
\label{tab:methods}
\setlength{\tabcolsep}{3pt}
\begin{tabular}{@{}llll@{}}
\toprule
Method & Mode & Evidence access & Judgment \\
\midrule
Direct & no-think & full fact table & model \\
Generic Agent & think & field-level tool & model \\
Schema Zero-shot & no-think & specification & verifier \\
Schema Few-shot & no-think & specification & verifier \\
\textbf{Ours SFT} & no-think & specification & verifier \\
\bottomrule
\end{tabular}
\end{table}

\textbf{Schema-based methods.} Zero-shot receives only the schema instructions, which list the 38 keys and ask for requirements outside them in a separate field. Few-shot adds three synthetic demonstrations that cover a broad range of keys. Ours SFT uses the Few-shot prompt after fine-tuning.

\textbf{Generic Agent.} The agent uses Qwen-Agent\footnote{\url{https://github.com/QwenLM/Qwen-Agent}}, in thinking mode (the model's default), with default execution settings and no budget overrides. It receives the clause, the field list, and one tool, \texttt{get\_document\_information(document\_type, field\_name)}, but no schema or verifier. It identifies requirements, fetches fields, reports checks (the prompt asks for one check per requirement), and gives a case decision. A logged lookup counts as retrieved evidence only if it returns the stored value of the requested field; an empty string counts only when the field is actually empty, not when a lookup fails.

\textbf{Direct.} Direct receives the clause and the full fact table in one non-thinking prompt. It returns PASS/FAIL for each key it examines and a case label. We call Direct and the Agent the model-based baselines.

\textbf{Gold-spec execution.} As a diagnostic reference, we also execute the gold encodings with the same verifier. It shows which errors remain when the specification matches the annotation, and thus separates generation errors from errors that the representation itself introduces. It is not an upper bound on the outcome-level metrics (Section~\ref{sec:rq1}); Section~\ref{sec:analysis} analyzes its disagreements with the human labels.

\textbf{Decoding.} 
Direct, Zero-shot, and Few-shot use the model card’s non-thinking setting (temperature 0.7, top-p 0.8), and the Agent uses its thinking setting (temperature 1.0, top-p 0.95); all methods use top-k 20 and presence penalty 1.5.\footnote{\url{https://huggingface.co/Qwen/Qwen3.5-9B}} These values are recommended for the released weights. SFT changes the weights and was trained to emit one target specification per clause, so we decode it greedily to keep that format stable. All other sampling parameters are those of Few-shot, so the two differ in decoding only in temperature.
AceSearcher \cite{xu2025acesearcher} uses the same split: it follows the official guidance for its Qwen3 baselines (temperature 0.7 in non-thinking and 0.6 in thinking mode) and decodes its own fine-tuned models at temperature 0. No decoding parameter was tuned on the evaluation cases.

\textbf{Comparability.} For the schema-based methods and the gold encodings, the verifier applies normalization and the CLEAN rule (Section~\ref{sec:exec}). Direct and the Agent receive the CLEAN rule in their prompts and compare values themselves. Since the systems also differ in evidence access, output format, and reasoning mode, we treat comparisons with the model-based baselines as system-level.

\subsection{Metrics}
\label{sec:metrics}

Figure~\ref{fig:layers} shows which observable layers each metric compares. For the schema-based methods, all metrics are computed after schemas on ontology keys that were placed in the out-of-ontology field are moved back (Section~\ref{sec:exec}).
\figLayers

\textbf{Detection} counts (document, key) occurrences with multiplicity. Let $g_c(d,k)$ be the number of schemas in $S(c)$ with document $d$ and key $k$, and $p_c(d,k)$ the predicted count. Then
\begin{gather*}
m=\sum_c\sum_{d,k}\min\{g_c(d,k),p_c(d,k)\},\\
\mathrm{AtomicRecall}=\frac{m}{\sum_c\sum_{d,k}g_c(d,k)},\\
\mathrm{AtomicPrecision}=\frac{m}{\sum_c\sum_{d,k}p_c(d,k)}.
\end{gather*}
For the Agent, predicted keys are the evidence fields of its checks; for Direct, they are its reported keys. Detection measures key coverage only; operators and values are checked in reconstruction. Unit Detection Recall is the share of gold units with at least one matched atomic schema. A predicted occurrence marks every gold unit that contains its (document, key), so this metric is lenient when units share a key; Atomic Recall counts occurrences one-to-one. Detection covers all 3,848 units.

\textbf{Reconstruction.} An exact unit match requires the correct document, key, operator, and expected value for every atomic schema, and the full logical structure. Expected values are compared as written, without normalization. Emitting only one alternative of \texttt{OR(BL.notify\_party EQUALS A, BL.notify\_party EQUALS B)}, or replacing OR with AND, is not exact, even if the outcome does not change. Exact Unit Recall is the share of gold units matched exactly. Perfect Cases is the share of cases whose predicted units match the gold units one-to-one, with none missing or extra. Both apply only to schema-based methods.

\textbf{Verification.} For schema-based methods, each predicted unit is executed as generated. It is then matched greedily and one-to-one to a gold unit of the same case that shares a (document, key) pair with it, preferring units that also agree in operator and expected value. Matching ignores outcomes and does not require exactness. If one gold unit was generated as several units, the extra units are still executed and enter the case decision. Checks of the model-based baselines are aligned by (document, key) counts, as in detection. Several Agent checks on one gold unit are combined by AND, the case-decision rule, and are not read as alternatives of an OR unit, because the prompt asks for one check per requirement.

Each evaluable gold unit then receives one terminal status: MISS if no check is aligned; UNGROUNDED if the Agent has not retrieved the required evidence (a field counts only if it is listed for the check and was successfully looked up); otherwise CORRECT or WRONG\_DECISION, by agreement with the human outcome. All outputs could be parsed. With $U_{eval}$ the 2,817 evaluable units and $U_A$, $U_G$, $U_C$ the addressed, grounded, and correct units among them,
\begin{gather*}
\mathrm{Addressed}=\frac{|U_A|}{|U_{eval}|},\quad
\mathrm{Grounding}=\frac{|U_G|}{|U_A|},\\
\mathrm{Judgment}=\frac{|U_C|}{|U_G|},\quad
\mathrm{UnitSuccess}=\frac{|U_C|}{|U_{eval}|}.
\end{gather*}
Judgment Accuracy is conditional on grounded units, which differ across methods; Unit Success keeps omissions and missing evidence in the denominator. Direct leaves no retrieval trace, so its grounding cannot be measured. We count its addressed units as grounded; its Grounding and Judgment Accuracy are then not comparable, so we report only its Unit Success (47.60\%, in the caption of Table~\ref{tab:main}).

\textbf{Case level.} Case Decision Accuracy compares a method's final label with the gold decision within the evaluated scope. Schema-based methods take the conjunction of the executed units. For the model-based baselines, we drop checks that concern only excluded requirements and take the conjunction of the rest; before this filtering, Direct's own label always equals the conjunction of its checks. A model-based check gives one outcome for all the fields it lists, so a check that mixes an excluded and an evaluable field cannot be reduced the way a composite gold unit is (Section~\ref{sec:eligibility}). We keep such a check, and its outcome stands for the evaluable unit: a failed Agent check on \texttt{endorsement} and \texttt{consignee} counts as FAIL for the consignee unit. False rejections are gold PASS cases judged FAIL; false acceptances are the reverse.

The \textbf{Complete Examination Rate} (CER) is the share of cases in which every evaluable gold unit is CORRECT and the final decision is correct. On FAIL cases, it therefore requires every evaluable discrepancy to be found, since a discrepancy missing from the single notice of Art.~16(c) cannot be relied on later. Like Unit Success, CER is outcome-level: an aligned check is CORRECT when it gives the human outcome, whether or not it is exact. A complete examination in this sense means that every evaluable requirement received the right outcome. Whether each check is also the intended one is measured by Exact Unit Recall and Perfect Cases. CER also ignores extra checks unless they change the decision; Perfect Cases and Atomic Precision cover them. Incompl.\ is the share of final-correct cases with at least one unsuccessful evaluable unit.

\section{Results}
\label{sec:results}

\begin{table}[t]
\centering\small
\caption{Unit verification on the 2,817 evaluable gold units (\%). Grounding is conditional on addressed units, and Judgment on grounded units. FAIL: gold FAIL units correctly identified, of 86. Direct, which has no retrieval step, reaches 47.60\% Unit Success. Gold spec.: executed gold encodings, a diagnostic reference (Section~\ref{sec:methods}).}
\label{tab:main}
\setlength{\tabcolsep}{2.5pt}
\begin{tabular}{@{}l
  S[table-format=3.2] S[table-format=3.2] S[table-format=2.2]
  S[table-format=2.2] S[table-format=2.0]@{}}
\toprule
Method & {Addr.} & {Ground.} & {Judg.} & {Unit Succ.} & {FAIL} \\
\midrule
Generic Agent & 73.62 & 91.37 & 80.53 & 54.17 & 28 \\
Schema Zero-shot & 87.40 & 100.00 & 74.49 & 65.10 & 38 \\
Schema Few-shot & 90.27 & 100.00 & 84.86 & 76.61 & 37 \\
\textbf{Ours SFT} & \bfseries 91.55 & 100.00 & \bfseries 89.96 & \bfseries 82.36 & \bfseries 39 \\
\midrule
\textit{Gold spec.} & \itshape 100.00 & \itshape 100.00 & \itshape 93.54 & \itshape 93.54 & \itshape 49 \\
\bottomrule
\end{tabular}
\end{table}

\begin{table}[t]
\centering\small
\caption{Case-level results within the evaluated scope (281 gold PASS, 64 gold FAIL). FR/FA: false rejections/acceptances; Acc.: Case Decision Accuracy; Bal.: balanced accuracy; Incompl.: share of final-correct cases with an unsuccessful evaluable unit.}
\label{tab:case}
\setlength{\tabcolsep}{3pt}
\begin{tabular}{@{}l
  S[table-format=3.0] S[table-format=2.0] S[table-format=2.2]
  S[table-format=2.2] S[table-format=2.2] S[table-format=3.2]@{}}
\toprule
Method & {FR} & {FA} & {Acc.} & {Bal.} & {CER} & {Incompl.} \\
\midrule
Always PASS & 0 & 64 & 81.45 & 50.00 & 0.00 & 100.00 \\
Always FAIL & 281 & 0 & 18.55 & 50.00 & 0.00 & 100.00 \\
\midrule
Direct & 43 & 58 & 70.72 & 47.04 & 0.00 & 100.00 \\
Generic Agent & 243 & 11 & 26.38 & 48.17 & 0.00 & 100.00 \\
Schema Zero-shot & 261 & 4 & 23.19 & 50.43 & 7.54 & 67.50 \\
Schema Few-shot & 201 & 5 & 40.29 & 60.33 & 25.51 & 36.69 \\
\textbf{Ours SFT} & 132 & 16 & 57.10 & \bfseries 64.01 & \bfseries 46.96 & \bfseries 17.77 \\
\midrule
\textit{Gold spec.} & \itshape 56 & \itshape 17 & \itshape 78.84 & \itshape 76.75 & \itshape 67.54 & \itshape 14.34 \\
\bottomrule
\end{tabular}
\end{table}

\begin{table}[t]
\centering\small
\caption{Detection and reconstruction (\%) over 4,493 atomic schemas and 3,848 gold units. Exact: Exact Unit Recall; Perfect: Perfect Cases. Few-shot and SFT share the same three-example prompt.}
\label{tab:reconstruction}
\setlength{\tabcolsep}{2.5pt}
\begin{tabular}{@{}l
  S[table-format=2.2] S[table-format=2.2] S[table-format=2.2]
  S[table-format=2.2] S[table-format=2.2]@{}}
\toprule
Method & {Atom.\,R} & {Atom.\,P} & {Unit Det.} & {Exact} & {Perfect} \\
\midrule
Direct & 44.25 & \bfseries 87.42 & 51.04 & {n/a} & {n/a} \\
Generic Agent & 65.35 & 62.88 & 70.84 & {n/a} & {n/a} \\
Schema Zero-shot & 78.19 & 81.89 & 87.71 & 57.64 & 0.00 \\
Schema Few-shot & 88.80 & 86.97 & 91.71 & 76.53 & 10.14 \\
\textbf{Ours SFT} & \bfseries 93.19 & 86.94 & \bfseries 93.32 & \bfseries 85.76 & \bfseries 27.83 \\
\bottomrule
\end{tabular}
\end{table}

\begin{table}[t]
\centering\small
\caption{Terminal statuses on 2,817 evaluable gold units.}
\label{tab:funnel}
\setlength{\tabcolsep}{4pt}
\begin{tabular}{@{}l
  S[table-format=4.0] S[table-format=4.0] S[table-format=4.0] S[table-format=4.0]@{}}
\toprule
Outcome & {Agent} & {Zero-shot} & {Few-shot} & {SFT} \\
\midrule
MISS & 743 & 355 & 274 & 238 \\
UNGROUNDED & 179 & 0 & 0 & 0 \\
WRONG\_DECISION & 369 & 628 & 385 & 259 \\
CORRECT & 1526 & 1834 & 2158 & 2320 \\
\bottomrule
\end{tabular}
\end{table}

\subsection{RQ1: Final decisions and complete examination}
\label{sec:rq1}

Correct final decisions often hide incomplete examination (Table~\ref{tab:case}). Every final-correct case of Direct (244) and of the Agent (91) contains an unsuccessful evaluable unit, as do 54 of 80 for Zero-shot, 51 of 139 for Few-shot, and 35 of 197 for SFT. Direct leaves out at least one evaluable unit in every case, partly by design: it reports one judgment per key and cannot address several units on the same key.

CER is 46.96\% (162 cases) for SFT, 25.51\% (88) for Few-shot, 7.54\% (26) for Zero-shot, and zero for both model-based baselines. Since all systems see the same cases, the differences can be tested in pairs: an exact McNemar test gives $p<10^{-5}$ for SFT against Few-shot and $p<10^{-8}$ for Few-shot against Zero-shot. The test accounts for case sampling but not for run-to-run variation. At least 13 of SFT's 162 completely examined cases are FAIL cases, because it decides only 149 PASS cases correctly; for Few-shot and Zero-shot, the corresponding numbers are at least 8 and 6.

Raw accuracy hides this. Always PASS scores highest (81.45\%) simply because 281 of the 345 cases are PASS, and Direct, the most accurate evaluated system (70.72\%), accepts 58 of the 64 discrepant cases; SFT comes next at 57.10\%. With balanced accuracy \cite{brodersen2010balanced}, which removes the majority-class advantage, SFT ranks first among the evaluated systems (64.01\%) and Few-shot second (60.33\%), while both model-based baselines fall below the 50\% of the constant rules. SFT is also the only evaluated system that decides more than half of each class correctly (149/281 PASS, 48/64 FAIL). Even so, no decision-level metric can tell whether a correct decision came from a complete examination. Of SFT's 16 false acceptances, 6 occur because a failing unit was never generated and 10 because a unit was misjudged.

A correct refusal can also be based on the wrong grounds. Of the 86 gold FAIL units, SFT correctly identifies 39, Zero-shot 38, Few-shot 37, and the Agent 28 (Table~\ref{tab:main}), and each of them forces a FAIL decision in its case. SFT correctly refuses 48 cases, so at least 9 of these refusals come only from other failing checks, such as false FAIL judgments or extra units. For the Agent, Zero-shot, and Few-shot, more than a third of the correct refusals are of this kind. 

Completing the examination of a discrepant case is hard even without generation errors. Executing the gold encodings gives 47 correct refusals, but only 8 of them examine every evaluable unit correctly. A correct refusal needs one failing unit, while a complete examination needs every unit right. In a case with one discrepancy and several satisfied requirements, finding the discrepancy is enough to refuse correctly, but a satisfied requirement that is wrongly failed, or a second discrepancy that is missed, leaves the examination incomplete. Gold execution produces 145 false FAILs across all cases (Section~\ref{sec:analysis}). SFT completes more FAIL cases than gold execution (at least 13 against 8) while finding fewer discrepancies (39 against 49), because CER is outcome-level: an inexact generated unit can pass a satisfied requirement on which the gold encoding wrongly fails. This is also why gold execution is a diagnostic reference here and cannot serve as an upper bound.

\subsection{RQ2: Specification-guided and model-based examination}
\label{sec:rq2}

SFT reaches 82.36\% Unit Success, against 54.17\% for the Agent (Table~\ref{tab:main}), and Few-shot already reaches 76.61\% without training. These figures need care. Since 96.95\% of the evaluable units are PASS, a system that addressed every unit and passed all of them would reach 96.95\% without finding a single discrepancy, and for this reason Table~\ref{tab:class} splits Unit Success by gold class. SFT's lead of 794 units over the Agent comes almost entirely from PASS units (783); overall, SFT has 505 fewer misses, 179 fewer ungrounded units, and 110 fewer wrong decisions (Table~\ref{tab:funnel}). Most of the gain thus comes from examining satisfied requirements and passing them correctly. On the 86 FAIL units, Unit Success rises from 32.56\% to 45.35\%, and the balanced mean keeps the ordering of Table~\ref{tab:main} (43.70\% for the Agent, 64.44\% for SFT). The PASS-unit gains matter for completeness: in a compliant case every evaluable unit is a PASS unit, and only a system that gets all of them right examines the case completely.

Direct has the lowest Unit Detection Recall (51.04\%) yet the highest Atomic Precision (87.42\%; Table~\ref{tab:reconstruction}). The keys it reports are usually required ones; it simply reports too few of them. Zero-shot beats the Agent in Unit Success (65.10\%) despite having the lowest Judgment Accuracy (74.49\%): it covers more units, although its units more often execute to the wrong outcome. Many of its units are inexact (Exact Unit Recall 57.64\%), but matching does not require exactness, and when most units are PASS, an inexact unit often still gives the right outcome. The Agent and Zero-shot reject most compliant cases (243 and 261 of 281). Under conjunction a single wrong FAIL is enough, and the Agent's mixed checks also carry failures over from excluded fields.

\begin{table}[t]
\centering\small
\caption{Unit Success (\%) on the 2,731 PASS and 86 FAIL evaluable units; Balanced is their unweighted mean. Derived from Tables~\ref{tab:main} and~\ref{tab:funnel}.}
\label{tab:class}
\setlength{\tabcolsep}{4pt}
\begin{tabular}{@{}l
  S[table-format=2.2] S[table-format=2.2] S[table-format=2.2]@{}}
\toprule
Method & {PASS} & {FAIL} & {Balanced} \\
\midrule
Generic Agent & 54.85 & 32.56 & 43.70 \\
Schema Zero-shot & 65.76 & 44.19 & 54.97 \\
Schema Few-shot & 77.66 & 43.02 & 60.34 \\
\textbf{Ours SFT} & \bfseries 83.52 & \bfseries 45.35 & \bfseries 64.44 \\
\midrule
\textit{Gold spec.} & \itshape 94.69 & \itshape 56.98 & \itshape 75.83 \\
\bottomrule
\end{tabular}
\end{table}

\subsection{RQ3: Fine-tuning on synthetic specifications}
\label{sec:rq3}

With the same three-example prompt, SFT raises Atomic Recall from 88.80\% to 93.19\% while Atomic Precision stays the same (86.94\% versus 86.97\%). The larger changes are in Exact Unit Recall, from 76.53\% to 85.76\%, and in Perfect Cases, from 10.14\% (35/345) to 27.83\% (96/345); Unit Detection Recall rises by only 1.61 points. What the fine-tuned parser gains, then, is mainly the ability to recover complete atomic schemas and their structure. Some of this may come from the encoding conventions that training targets share with the annotations (Section~\ref{sec:synthesis}), such as the OR encoding of CLEAN; they can favor SFT on Exact Unit Recall and Perfect Cases, which compare with the encodings. Unit Success and CER compare outcomes with the human labels, and they rise as well.

Verification follows the same order. Compared with Few-shot, SFT makes 126 fewer wrong decisions and 36 fewer misses (162 more CORRECT units); Few-shot in turn makes 243 fewer wrong decisions and 81 fewer misses than Zero-shot. No schema-based run has an UNGROUNDED unit, because the verifier fetches every field that a schema names and the fact store holds every key. As in the comparison with the Agent, the advantage is on satisfied requirements: 160 of SFT's 162 additional CORRECT units are PASS units (Table~\ref{tab:class}). SFT cuts false rejections from 201 to 132 but raises false acceptances from 5 to 16. On discrepancies, the three schema-based configurations identify 37 to 39 FAIL units, a difference that 86 units and one run per configuration cannot resolve. Few-shot and SFT also differ in decoding (Section~\ref{sec:methods}), so the comparison is between two configurations and does not isolate the effect of training.

\section{Discussion}
\label{sec:analysis}

\textbf{What the representation allows.} To see what the representation itself allows, we executed the gold encodings with the same verifier. They match the human outcomes on 2,635 of 2,817 units (93.54\%), 315 more than SFT, and identify 49 of the 86 FAIL units. At the case level they make 56 false rejections and 17 false acceptances, which puts their accuracy (78.84\%) below that of Always PASS; their CER is 67.54\% (Table~\ref{tab:case}). Since gold execution misses no unit, its 182 disagreements with the human outcomes split into 145 false FAILs on satisfied requirements and 37 false PASSes on annotated discrepancies. Four in five of these errors flag a requirement for review, and only one in five lets a discrepancy through. Two finance-domain experts who had not taken part in the annotation reviewed the 182 disagreements and found no label or encoding error. Here, ``no encoding error'' means that each gold encoding follows the annotation conventions for its key; it does not mean that the six operators can verify every requirement on that key faithfully. The disagreements run in both directions. The 145 false FAILs come from comparisons stricter than examination practice: documents add data that do not conflict, write the same data differently, or vary within UCP tolerances (Art.~30(b)), and some requirements, such as a six-digit HS code, need format operators. The 37 false PASSes come from substring tests: CONTAINS passes a value that includes the required text even when the rest of it changes the meaning. Tightening or loosening the operators would trade one error for the other; what is needed is a comparison that judges whether data conflict, as Art.~14(d) does. Because these errors already occur with the gold encodings, a better parser alone cannot remove them.

\textbf{Error types and the examiner's role.} With only 86 FAIL units, most wrong decisions are false FAILs; for SFT, over 80\% of its 259 wrong decisions reject a satisfied requirement. Under conjunction a single wrong FAIL, possibly from an extra unit, is enough to reject a compliant case, and SFT rejects 132 of the 281 compliant cases (46.98\%) while accepting 16 of the 64 discrepant ones (25.00\%). The two errors have different costs: a false rejection costs examiner time, while a false acceptance can make a bank honour a discrepant presentation. At these rates the system cannot run on its own. A better fit is the hybrid AI--human examination recommended for LC practice \cite{khalil2025roadmap}, where every rejection reaches the examiner together with the units that caused it and the evidence they read. The trace cannot show the 238 units that SFT missed, though, and whether checking its flags saves examiners time compared with examining unaided is a question for a study with examiners.

\section{Limitations}
\label{sec:limitations}

Our study has several limits. It models only BL and IV requirements in field 46A; 378 requirements (7.76\%) fall outside the fixed ontology, and the parser's separate list of such requirements is not evaluated; verification covers 2,817 of 3,848 units. Comparisons with the model-based baselines are system-level. Each configuration was run once, which leaves run-to-run variation unmeasured, and Few-shot and SFT also differ in decoding. With 86 FAIL units in 64 cases, conclusions about discrepancy detection draw on few events. All cases come from one institution. The labels come from three in-house annotators who cross-reviewed one another's work; without independent double annotation, inter-annotator agreement is not available. Encoding conventions shared by templates and annotations may favor the fine-tuned parser on exact-match metrics. Finally, the real cases and the templates derived from confidential guidance cannot be released: others can generate training pairs with the released code and public-source templates, but cannot reproduce the reported scores.

\section{Conclusion}
\label{sec:conclusion}

On 345 real LC cases, final-decision accuracy says little about whether the examination was complete. A constant PASS rule has the highest raw accuracy while examining nothing, and correct refusals can be based on checks other than the annotated discrepancies. Evaluating requirement by requirement exposes this gap, and executable specifications make it measurable. They also show where errors come from. Finding discrepancies is still the weak point, and generation is only part of the reason: SFT finds 39 of the 86 annotated discrepancies, and even the gold encodings, executed with the same verifier, find only 49. Such specifications can be generated without confidential records: a parser trained only on synthetic pairs completely examines 46.96\% of the cases within the evaluated scope, 1.8 times as many as few-shot prompting under its recommended sampling setting, and reaches a class-balanced Unit Success of 64.44\%, against 43.70\% for a generic agent that examines no case completely. We see three next steps. Format operators and a comparison closer to Art.~14(d) would remove errors that gold execution exposes. A check of the clause against its specification could flag omissions that the trace cannot show. Extending the study to other documents, field 47A, and document images would test the approach end to end.

\begin{acks}
This research was supported by Hana Financial Group. It also used
datasets from `The Open AI Dataset Project (AI-Hub, S.~Korea)'. All
data information can be accessed through `AI-Hub (\url{www.aihub.or.kr})'.
\end{acks}

\end{document}